\documentclass[runningheads]{llncs}
\usepackage[T1]{fontenc}
\usepackage{graphicx}
\usepackage{booktabs}
\usepackage{amsfonts}
\usepackage{amsmath}
\usepackage{subcaption}
\usepackage{float}
\usepackage{url}
\usepackage{authblk}

\title{AST-n: A Fast Sampling Approach for Low-Dose CT Reconstruction using Diffusion Models}%

\titlerunning{AST-n}
\author{Tomás de la Sotta\inst{1} \orcidID{0009-0001-5608-7657} \and
José Manuel Saavedra\inst{1}\orcidID{0000-0002-9644-5164} \and
Héctor Henríquez\inst{1}\orcidID{0000-0003-1944-5336} \and
Violeta Chang\inst{2}\orcidID{0000-0002-9044-4849}\and
Aline Xavier\inst{2}\orcidID{0000-0001-6793-341X}}

\institute{Universidad de los Andes, Chile \and  Universidad de Santiago de Chile}
 
\begin{document}

\maketitle              
\begin{abstract}
Low-dose CT (LDCT) protocols reduce radiation exposure but increase image noise, compromising diagnostic confidence. Diffusion-based generative models have shown promise for LDCT denoising by learning image priors and performing iterative refinement. In this work, we introduce AST-n, an accelerated inference framework that initiates reverse diffusion from intermediate noise levels, and integrate high-order ODE solvers within conditioned models to further reduce sampling steps. We evaluate two acceleration paradigms—AST-n sampling and standard scheduling with high-order solvers—on the Low Dose CT Grand Challenge dataset, covering head, abdominal, and chest scans at 10–25 \% of standard dose. Conditioned models using only 25 steps (AST-25) achieve peak signal-to-noise ratio (PSNR) above 38 dB and structural similarity index (SSIM) above 0.95, closely matching standard baselines while cutting inference time from $\sim$16 s to under 1 s per slice. Unconditional sampling suffers substantial quality loss, underscoring the necessity of conditioning. We also assess DDIM inversion, which yields marginal PSNR gains at the cost of doubling inference time, limiting its clinical practicality. Our results demonstrate that AST-n with high-order samplers enables rapid LDCT reconstruction without significant loss of image fidelity, advancing the feasibility of diffusion-based methods in clinical workflows.

\end{abstract}
%
%
%
%

\section{Introduction}
\label{sec:introduction}

Radiological imaging has become one of the most essential tools for diagnosing pathologies, monitoring disease progression, and guiding therapeutic interventions. Among these, ionizing radiation-based diagnostic modalities constitute one of the main sources of radiation exposure in the general population, accounting for approximately 50\% of total exposure to ionizing radiation, including both natural and anthropogenic sources~\cite{laura_ncrp_2015}. In recent decades, the number of radiological examinations has increased considerably, contributing to faster and more accurate diagnoses and reducing the reliance on invasive, high-risk procedures.

Computed tomography (CT) has emerged as one of the most widely used modalities due to its excellent diagnostic performance and rapid acquisition. However, it is also considered the single largest contributor to medical radiation exposure~\cite{brenner_computed_2007}. For example, while a chest X-ray delivers an effective dose of approximately 0.1 mSv—roughly equivalent to 10 days of natural background exposure—a CT scan typically delivers between 1.5 and 15.4 mSv, depending on the anatomical region, protocol complexity, and the number of acquisition phases~\cite{noauthor_radiation_nodate}. In specific cases, doses can reach up to 37 mSv. The biological effects of ionizing radiation are cumulative and may include damage to genetic material, increasing the risk of radiation-induced neoplasms. Although estimating this risk is complex, Brenner et al.~\cite{brenner_computed_2007} suggest that between 1.5\% and 2.0\% of neoplasms may be attributable to current CT usage, with risks being notably higher for younger individuals. Consequently, minimizing dose exposure while maintaining diagnostic quality remains a pressing challenge, particularly in pediatric and young adult populations.

To address this, low-dose CT (LDCT) protocols attempt to reduce radiation exposure by decreasing X-ray intensity. However, this reduction often results in significant image noise, adversely affecting diagnostic reliability. Although traditional denoising and iterative reconstruction methods can partially mitigate these effects, their performance tends to deteriorate under extreme low-dose conditions. In recent years, deep learning has emerged as a powerful tool for medical image enhancement, including LDCT reconstruction. Multiple neural-network-based approaches have demonstrated strong results~\cite{chen_low-dose_2017,xia_low-dose_2022,yang_low-dose_2018}, particularly those employing diffusion-based generative models, which leverage learned image priors and iterative refinement processes~\cite{huang_one_2022,gao_cocodiff_2022,gao_corediff_2023}.

Despite the rapid progress in this field, several key limitations remain unaddressed. Although diffusion-based models have achieved excellent reconstruction quality, they often require long inference times, which are frequently incompatible with the time-sensitive nature of CT examinations. In clinical settings, CT is valued precisely for its speed, and introducing reconstruction latencies of several seconds per volume can hinder its practical deployment. Furthermore, many existing models are optimized to predict residual noise rather than directly reconstruct the image content, a strategy that has been shown to be suboptimal in restoration tasks~\cite{niu_acdmsr_2023}.

In this study, we explore the use of existing diffusion-based generative models for low-dose CT reconstruction by incorporating two complementary strategies to reduce inference time and enhance clinical applicability. First, we implement a low-dose conditioning scheme based on slice-level input concatenation, which provides anatomical context to guide the generative process. Second, we adopt AST-n, a timestep reduction strategy that initiates the reverse diffusion from intermediate noise levels instead of pure Gaussian noise, leading to a significant reduction in inference time.

Our main contribution lies in the systematic integration of AST-n sampling into an existing conditioned generative diffusion framework for LDCT reconstruction—without modifying the model architecture or requiring additional training. This enables previously trained models to be retrofitted with accelerated inference capabilities at minimal computational cost, enhancing their usability with almost no degradation in the model’s performance. By substantially reducing the number of sampling steps, our approach lowers inference time by an order of magnitude, moving diffusion-based reconstruction closer to real-time applicability. Clinically, this not only increases the feasibility of high-quality LDCT imaging in time-critical scenarios—ultimately contributing to safer diagnostic practices by making dose-reduction strategies more accessible in healthcare—but also enables faster image availability and improved workflow efficiency, especially in contexts where diagnostic speed is essential.

For clarity and ease of reading, the remainder of this paper is structured as follows. Section~2 reviews related work on LDCT reconstruction using generative models. Section~3 presents our proposed approach, including dataset details, model architecture, and acceleration strategies. Section~4 reports and discusses the experimental results. Finally, Section~5 concludes the paper and outlines directions for future research.

\section{Related Work}
\label{sec:related}

The generation of high-quality images from low-dose CT (LDCT) acquisitions has become a prominent challenge in radiological imaging. While reduced radiation protocols are crucial for minimizing patient risk, particularly in repeated scans and vulnerable populations, they significantly degrade image quality due to low signal-to-noise ratios. To address this, numerous deep learning approaches have been proposed in recent years, aiming to reconstruct full-dose-equivalent images from noisy LDCT inputs.

One of the earliest efforts in this direction was proposed by Chen et al.~\cite{chen_low-dose_2017}, who used a convolutional neural network to directly map low-dose images to their full-dose counterparts. This line of work laid the foundation for supervised denoising in CT using image-to-image learning. More recent approaches have explored diffusion-based generative models, leveraging their strong prior modeling capabilities and iterative refinement properties.

Xia et al.~\cite{xia_low-dose_2022} applied Denoising Diffusion Probabilistic Models (DDPMs) to LDCT reconstruction and introduced an acceleration scheme based on modeling the reverse process with ordinary differential equations (ODEs), following the formulation of Lu et al.~\cite{lu_dpm-solver_2022}. Their model, conditioned on the low-dose image, achieved substantial improvements in reconstruction quality while reducing inference time by up to 20×.

In a complementary approach, Huang et al.~\cite{huang_one_2022} addressed the problem in the sinogram domain. They trained a diffusion model to infer the ideal sinogram from low-dose projections, using low-rank Hankel priors to regularize the reconstruction. This strategy allowed them to operate in the raw projection space before applying traditional image reconstruction techniques.

Gao and Shan~\cite{gao_cocodiff_2022} proposed CoCoDiff, a context-conditioned diffusion model that learns the residual between low-dose and full-dose images. During training, the model estimates noise levels for each timestep, guided by adjacent slice information to preserve structural consistency. Inference proceeds by reversing the diffusion process conditioned on both the low-dose image and spatial context, yielding high-quality outputs with reduced oversharpening.

More recently, CoreDiff~\cite{gao_corediff_2023} introduced an alternative formulation where the forward process begins from the low-dose image rather than Gaussian noise. This significantly shortens the sampling trajectory by assuming that the input already contains partial information about the final image. A mean-preserving degradation operator and a contextually error-modulated restoration network (CLEAR-Net) are used to simulate and correct for physical degradation and accumulated errors, respectively. This model was especially effective in ultra-low-dose scenarios, achieving competitive results with as few as 10 sampling steps.

While these methods demonstrate excellent quantitative results using standard metrics such as PSNR and SSIM, several challenges remain. Many models prioritize residual prediction rather than full content reconstruction, which has been shown to limit performance~\cite{niu_acdmsr_2023}. Evaluation remains mostly focused on pixel-level similarity, with limited consideration of diagnostic utility. Furthermore, real-world datasets are scarce, with most benchmarks relying on simulated phantoms that fail to capture the true distribution of clinical noise. Lastly, although generative diffusion models offer strong reconstruction performance, their high inference times remain a barrier to clinical integration in time-sensitive applications like CT.

\section{Proposal}
\label{sec:proposal}

\subsection{Dataset}

The dataset used in this study is the Low Dose CT Grand Challenge, which includes 295 clinical CT volumes across three anatomical regions: 99 head, 97 abdomen, and 99 chest. Low-dose acquisition was simulated by injecting Poisson-distributed noise into the sinograms, corresponding to 25\% of standard radiation dose for head and abdominal scans, and 10\% for chest scans. All images were provided in DICOM format and converted to floating-point arrays, then normalized to the \([0,1]\) range using fixed intensity bounds \([-1024, 3072]\).

\subsection{Diffusion Process and Conditioning}

We adopt the standard formulation of denoising diffusion probabilistic models (DDPMs), where a forward Markov process gradually adds Gaussian noise to a data sample \( x_0 \sim q(x_0) \) over \( T \) discrete steps. This process is defined as:
\begin{equation}
    q(x_t \mid x_{t-1}) = \mathcal{N}(x_t; \sqrt{1 - \beta_t} \, x_{t-1}, \beta_t \, I),
\end{equation}
with \(\beta_t \in (0, 1)\) controlling the noise schedule. By composing the transitions, we obtain the marginal:
\begin{equation}
    q(x_t \mid x_0) = \mathcal{N}(x_t; \sqrt{\bar{\alpha}_t} \, x_0, (1 - \bar{\alpha}_t) \, I),
\end{equation}
where \(\bar{\alpha}_t = \prod_{s=1}^t (1 - \beta_s)\) is the cumulative product of noise decay factors.

During training, the model learns to estimate the noise component \(\epsilon\) from a noisy sample \( x_t \), the timestep \( t \), and a conditioning signal \( c \), here given by the corresponding low-dose CT image. The objective function is the expected squared error:
\begin{equation}
    L(\theta) = \mathbb{E}_{x_0, \epsilon, t} \left[ \left\| \epsilon_\theta(x_t, t, c) - \epsilon \right\|_2^2 \right],
\end{equation}
where \( \epsilon \sim \mathcal{N}(0, I) \). Conditioning is incorporated at each denoising step to guide the reconstruction process.

\subsection{DDIM Inversion}

To assess whether the model can traverse the latent trajectory both forward and backward, we evaluate deterministic inversion using DDIM as proposed in \cite{mokady_null-text_2022}. Starting from a low-dose image \( x_0 \), we compute an approximate latent \( x_T \) by applying the DDIM inversion rule for \( T = 1000 \) steps:
\begin{equation}
    x_{t+1} = \sqrt{\bar{\alpha}_{t+1}} \, x_0 + \sqrt{1 - \bar{\alpha}_{t+1} - \sigma_{t+1}^2} \cdot \frac{x_t - \sqrt{\bar{\alpha}_t} \, x_0}{\sqrt{1 - \bar{\alpha}_t}}, \quad \sigma_t = 0.
\end{equation}
The resulting latent \( x_T \) is then used as the starting point for a standard reverse sampling process. This setup evaluates whether the model can map a real input to a plausible point in the latent space and reconstruct it without loss, thereby testing the consistency of the learned generative trajectory.

\subsection{AST-n Sampling Strategy}

We evaluate an acceleration strategy termed AST-n (Accelerated Sampling from Time-step~\(n\)), where generation begins from an intermediate latent \( x_t \) instead of from pure Gaussian noise \( x_T \sim \mathcal{N}(0, I) \). The value of \( t \) is chosen such that \( t \ll T \), typically within the range \( t \in \{10, 25, 50, 100, 150, 500\} \), with \( T = 1000 \) being the total number of diffusion steps used during training.

Given a clean image \( x_0 \), the latent \( x_t \) is analytically computed using the closed-form of the forward process:
\begin{equation}
    x_t = \sqrt{\bar{\alpha}_t} \, x_0 + \sqrt{1 - \bar{\alpha}_t} \, \epsilon, \quad \epsilon \sim \mathcal{N}(0, I),
\end{equation}
where \( \bar{\alpha}_t = \prod_{s=1}^t (1 - \beta_s) \). The reverse process is then executed from \( x_t \) to reconstruct \( x_0 \), using the learned transitions:
\begin{equation}
    x_{t-1}, x_{t-2}, \dots, x_0 \sim p_\theta(x_{s-1} \mid x_s), \quad \text{for } s = t, \dots, 1.
\end{equation}

This strategy is theoretically grounded in the DDPM framework, where the marginal distribution \( q(x_t \mid x_0) \) remains Gaussian for all \( t \in [0, T] \). As such, starting the generative process from an intermediate point \( x_t \) does not violate the statistical assumptions under which the model was trained. By selecting \( t \ll T \), AST-n allows significant reduction in inference steps while maintaining the validity of the diffusion process. This approach enables fast image reconstruction through a shortened denoising trajectory, with minimal loss in fidelity compared to full-step sampling.

\subsection{Sampling Techniques}

To explore the trade-off between inference time and reconstruction quality, we evaluate several sampling strategies compatible with DDPMs. Each model is tested under two regimes. The first corresponds to full-noise sampling, where generation starts from a standard Gaussian latent \( x_T \sim \mathcal{N}(0, I) \) and proceeds for a reduced number of steps \( N \in \{1000, 500, 150, 50, 25, 10\} \). The second corresponds to AST-n, where generation begins from \( x_t \) with \( t = N \), and the reverse process is executed for the same number of steps.

In both regimes, we evaluate solvers that approximate the reverse process more efficiently than the original DDPM formulation. These include DDIM~\cite{song_denoising_2022}, which uses a deterministic, non-Markovian trajectory; DPMSolver~\cite{lu_dpm-solver_2022}, which leverages ODE discretization; and UniPC~\cite{zhao_unipc_2023}, a high-order predictor-corrector scheme designed for fast inference.

\subsection{Evaluation Protocol}

Reconstruction quality is assessed using peak signal-to-noise ratio (PSNR), structural similarity index (SSIM), and root mean squared error (RMSE). Inference efficiency is measured as the average reconstruction time per batch. All experiments are conducted on a held-out test set, and metrics are reported separately for each experimental configuration.

\section{Experimental Results and Discussion}
\label{sec:results}

We evaluated multiple sampling strategies over conditioned diffusion models for LDCT reconstruction. Table~\ref{tab:selected_results} summarizes PSNR, RMSE, SSIM and reconstruction time for the standard full‐schedule run (DDPM@1000), five samplers at 150 steps (DDPM@150, DDIM@150, DPMSolver-1@150, DPMSolver-2@150, DPMSolver++@150) and their AST-150 variants. AST-150 skips the earliest denoising steps—where conditioning is weakest—and concentrates computation on the most critical refinement phases. Both acceleration methods yield up to an 6× speed-up while preserving, or even enhancing, image quality. Notably, all evaluated configurations perform similarly to our own reimplementation of the methods proposed by Xia et al.~\cite{xia_low-dose_2022} with the proposed dataset, highlighting the effectiveness of accelerated and conditioned sampling strategies. For baseline comparison, a standard conditioned DDPM model is included in the table results.

\begin{table}[ht!]
\centering
\caption{PSNR, RMSE, SSIM and reconstruction time for Full‐Schedule and AST‐150. Each cell shows the comparison of results between the use of DDIM inversion (left) \cite{mokady_null-text_2022} and the standard DDPM \cite{ho_denoising_2020} noise addition strategy (right).}
\label{tab:selected_results}
\resizebox{\linewidth}{!}{
\begin{tabular}{lcccc}
\toprule
\textbf{Model}          & \textbf{PSNR (dB)} & \textbf{RMSE}        & \textbf{SSIM}     & \textbf{Time (s)}    \\
\midrule
DDPM@1000            & 38.788/38.721   & 27.785/27.779 & 0.973/0.973 & 45.144/16.382 \\

\midrule

DDPM@150             & 26.594/26.601   & 26.594/26.601 & 0.975/0.975 & 6.665/2.427 \\
DDIM@150             & 30.213/41.150   & 126.384/29.188 & 0.781/0.976 & 5.940/2.425   \\
Solver-1@150         & 39.758/39.757     & 24.884/24.884 & 0.978/0.978 & 6.598/2.416   \\
Solver-2@150         & 39.652/39.643     & 25.219/25.217 & 0.978/0.978 & 6.070/2.407   \\
Solver++@150         & 39.753/39.740     & 24.895/24.893 & 0.978/0.978 & 6.656/2.397   \\

\midrule

DDPM (AST-150)       & 39.070/38.778   & 26.901/27.766 & 0.975/0.973 & 6.578/2.433   \\
DDIM (AST-150)       & 39.085/39.272   & 27.030/26.461 & 0.972/0.974 & 6.264/2.397   \\
Solver-1 (AST-150)   & 39.745/33.324   & 24.881/52.088 & 0.978/0.894 & 6.647/2.417   \\
Solver-2 (AST-150)   & 39.745/33.233   & 24.888/52.651 & 0.978/0.892 & 6.605/2.411   \\
Solver++ (AST-150)   & 39.732/38.724   & 24.928/27.866 & 0.978/0.973 & 6.617/2.416   \\
\bottomrule
\end{tabular}
}
\end{table}

Figure~1 illustrates the exceptional stability and efficiency of AST-\(n\) acceleration compared to standard step-skipping. In Fig.~1(a), PSNR curves for all samplers under AST-\(n\) remain essentially flat—varying by less than 0.9dB  between 25 and 1000 steps—while SSIM and RMSE show equally minimal fluctuations, confirming negligible quality loss when skipping early denoising. By contrast, standard scheduling in Fig.~1(b) exhibits a pronounced PSNR drop below 150 steps and only recovers at high counts. At the same time, average reconstruction time per batch decreases almost linearly with \(n\), from approximately 45s at 1000 steps down to under 2.5s at \(n=150\) and under 0.4s at \(n=10\). Together, these results suggest that, under strong anatomical conditioning in the denoising pipeline—where the input image already encodes rich structural detail—the earliest denoising steps have only a limited impact on final image fidelity, which highlights AST-n as a robust and highly efficient acceleration strategy for LDCT reconstruction.

\begin{figure}[ht!]
\centering
\begin{subfigure}[t]{0.48\textwidth}
\centering
\includegraphics[width=\textwidth]{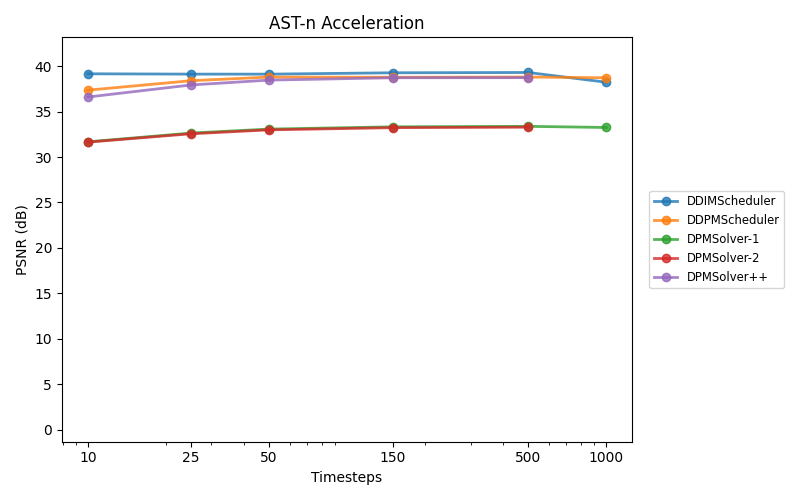}
\caption{AST-n Accelerated Models}
\end{subfigure}
\hfill
\begin{subfigure}[t]{0.48\textwidth}
\centering
\includegraphics[width=\textwidth]{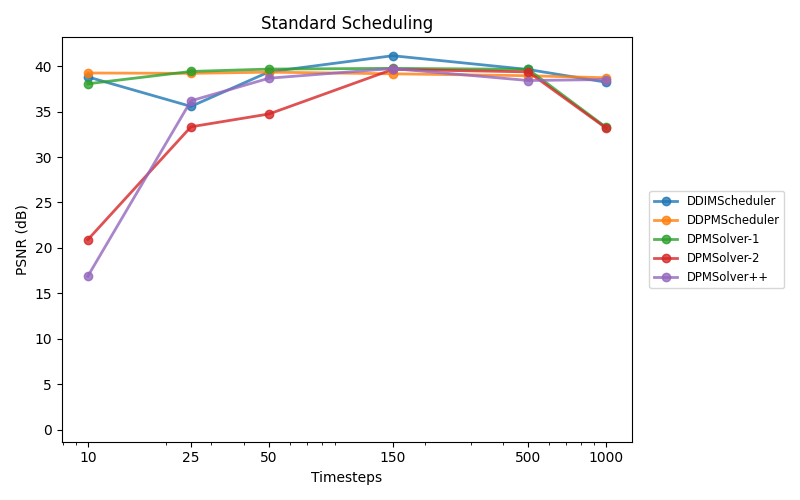}
\caption{Standard Scheduling}
\end{subfigure}
\caption{Comparison of PSNR values using AST-\textit{n} and standard step-skipping acceleration methods.}
\label{fig:eq-vs-last-psnr}
\end{figure}

\begin{figure}[ht!]
    \centering
    \includegraphics[width=\linewidth]{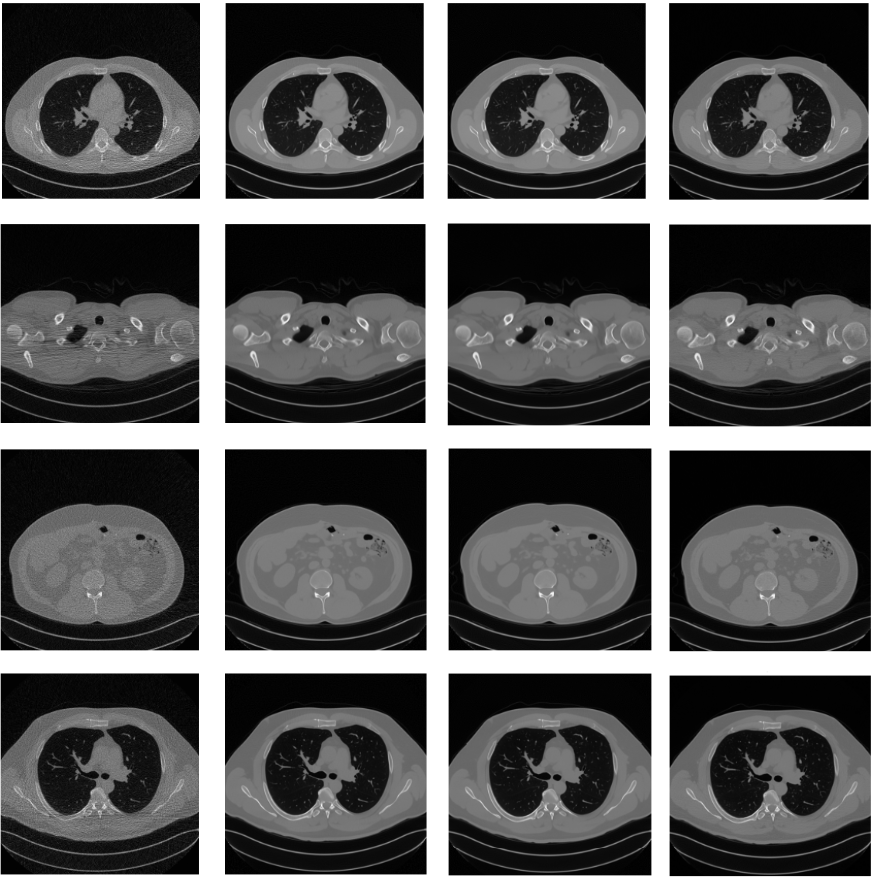}
    \caption{AST-n strategy over DDIM scheduler. From left to right, LDCT image, AST-10, AST-150, full schedule (1000 steps)}
    \label{fig:astn}
\end{figure}

\section{Conclusions}
\label{sec:conclusions}
Experimental results demonstrate that AST-\(n\) acceleration—initiating reverse diffusion from intermediate noise levels—yields reconstruction fidelity equivalent to full‐schedule baselines while reducing inference latency by more than an order of magnitude. Specifically, AST-150 paired with DDIM and DPMSolver models achieves PSNR \(\ge38\)dB and SSIM \(\approx0.97\) in under 2.5s, compared to 45s for a 1000-step DDPM run. AST-25 configurations further validate aggressive step reduction, achieving inference times below 0.4s while incurring only minor quality losses—PSNR drops remain under 3dB at 10 steps and under 0.9dB at 25 steps. The marginal quality gains offered by DDIM inversion do not justify its computational cost in practical LDCT scenarios, almost duplicating the reconstruction time for each image. Future research will investigate adaptive noise schedules and learned inversion schemes to improve robustness at ultra-low step counts and enable real-time clinical deployment.

\bibliographystyle{splncs04}
\bibliography{references}

\begin{thebibliography}{10}
\providecommand{\url}[1]{\texttt{#1}}
\providecommand{\urlprefix}{URL }
\providecommand{\doi}[1]{https://doi.org/#1}

\bibitem{noauthor_radiation_nodate}
Radiation {Dose} to {Adults} {From} {Common} {Imaging} {Examinations}. Tech. rep., American College of Radiology, \url{https://www.acr.org/-/media/ACR/Files/Radiology-Safety/Radiation-Safety/Dose-Reference-Card.pdf}

\bibitem{brenner_computed_2007}
Brenner, D.J., Hall, E.J.: Computed {Tomography} — {An} {Increasing} {Source} of {Radiation} {Exposure}. New England Journal of Medicine  \textbf{357}(22),  2277--2284 (Nov 2007). \doi{10.1056/NEJMra072149}

\bibitem{chen_low-dose_2017}
Chen, H., Zhang, Y., Zhang, W., Liao, P., Li, K., Zhou, J., Wang, G.: Low-dose {CT} via convolutional neural network. Biomedical Optics Express  \textbf{8}(2), ~679 (Feb 2017). \doi{10.1364/BOE.8.000679}

\bibitem{gao_corediff_2023}
Gao, Q., Li, Z., Zhang, J., Zhang, Y., Shan, H.: {CoreDiff}: {Contextual} {Error}-{Modulated} {Generalized} {Diffusion} {Model} for {Low}-{Dose} {CT} {Denoising} and {Generalization}. IEEE Transactions on Medical Imaging pp.~1--1 (2023). \doi{10.1109/TMI.2023.3320812}

\bibitem{gao_cocodiff_2022}
Gao, Q., Shan, H.: {CoCoDiff}: a contextual conditional diffusion model for low-dose {CT} image denoising. In: Müller, B., Wang, G. (eds.) Developments in {X}-{Ray} {Tomography} {XIV}. p.~16. SPIE, San Diego, United States (Oct 2022). \doi{10.1117/12.2634939}

\bibitem{ho_denoising_2020}
Ho, J., Jain, A., Abbeel, P.: Denoising diffusion probabilistic models. In: Larochelle, H., Ranzato, M., Hadsell, R., Balcan, M., Lin, H. (eds.) Advances in Neural Information Processing Systems 33: Annual Conference on Neural Information Processing Systems 2020, NeurIPS 2020, December 6-12, 2020, virtual (2020)

\bibitem{huang_one_2022}
Huang, B., Zhang, L., Lu, S., Lin, B., Wu, W., Liu, Q.: One {Sample} {Diffusion} {Model} in {Projection} {Domain} for {Low}-{Dose} {CT} {Imaging}  (2022). \doi{10.48550/ARXIV.2212.03630}, publisher: arXiv

\bibitem{laura_ncrp_2015}
{Laura}: {NCRP} {Report} 160 - {NCRP} {\textbar} {Bethesda}, {MD} (Jun 2015), \url{https://ncrponline.org/publications/reports/ncrp-report-160-2/}

\bibitem{lu_dpm-solver_2022}
Lu, C., Zhou, Y., Bao, F., Chen, J., Li, C., Zhu, J.: Dpm-solver: {A} fast {ODE} solver for diffusion probabilistic model sampling in around 10 steps. In: Koyejo, S., Mohamed, S., Agarwal, A., Belgrave, D., Cho, K., Oh, A. (eds.) Advances in Neural Information Processing Systems 35: Annual Conference on Neural Information Processing Systems 2022, NeurIPS 2022, New Orleans, LA, USA, November 28 - December 9, 2022 (2022)

\bibitem{mokady_null-text_2022}
Mokady, R., Hertz, A., Aberman, K., Pritch, Y., Cohen{-}Or, D.: Null-text inversion for editing real images using guided diffusion models. In: {IEEE/CVF} Conference on Computer Vision and Pattern Recognition, {CVPR} 2023, Vancouver, BC, Canada, June 17-24, 2023. pp. 6038--6047. {IEEE} (2023). \doi{10.1109/CVPR52729.2023.00585}, \url{https://doi.org/10.1109/CVPR52729.2023.00585}

\bibitem{niu_acdmsr_2023}
Niu, A., Pham, T.X., Zhang, K., Sun, J., Zhu, Y., Yan, Q., Kweon, I.S., Zhang, Y.: {ACDMSR:} accelerated conditional diffusion models for single image super-resolution. {IEEE} Trans. Broadcast.  \textbf{70}(2),  492--504 (2024). \doi{10.1109/TBC.2024.3374122}

\bibitem{song_denoising_2022}
Song, J., Meng, C., Ermon, S.: Denoising {Diffusion} {Implicit} {Models} (Oct 2022), \url{http://arxiv.org/abs/2010.02502}, arXiv:2010.02502 [cs]

\bibitem{xia_low-dose_2022}
Xia, W., Lyu, Q., Wang, G.: Low-{Dose} {CT} {Using} {Denoising} {Diffusion} {Probabilistic} {Model} for 20\${\textbackslash}times\$ {Speedup}  (2022). \doi{10.48550/ARXIV.2209.15136}, publisher: arXiv

\bibitem{yang_low-dose_2018}
Yang, Q., Yan, P., Zhang, Y., Yu, H., Shi, Y., Mou, X., Kalra, M.K., Zhang, Y., Sun, L., Wang, G.: Low-{Dose} {CT} {Image} {Denoising} {Using} a {Generative} {Adversarial} {Network} {With} {Wasserstein} {Distance} and {Perceptual} {Loss}. IEEE Transactions on Medical Imaging  \textbf{37}(6),  1348--1357 (Jun 2018). \doi{10.1109/TMI.2018.2827462}

\bibitem{zhao_unipc_2023}
Zhao, W., Bai, L., Rao, Y., Zhou, J., Lu, J.: {UniPC}: {A} {Unified} {Predictor}-{Corrector} {Framework} for {Fast} {Sampling} of {Diffusion} {Models} (Oct 2023). \doi{10.48550/arXiv.2302.04867}, \url{http://arxiv.org/abs/2302.04867}, arXiv:2302.04867 [cs]

\end{thebibliography}
\end{document}